\pdfoutput=1

\documentclass[11pt]{article}
\usepackage[]{emnlp2021}
\usepackage{times,latexsym}
\usepackage{url}
\usepackage[T1]{fontenc}
\usepackage[utf8]{inputenc}
\usepackage{algorithm2e}
\usepackage{tabularx}
\newcolumntype{Y}{>{\centering\arraybackslash}X}
\usepackage{booktabs}

\usepackage{microtype}
\usepackage{graphicx}
\usepackage{lscape} 
\usepackage{xcolor}
\usepackage{amssymb}
\usepackage{array}
\usepackage{subcaption} 
\usepackage{enumitem}
\usepackage[capitalise]{cleveref}
\usepackage{pmboxdraw}

\usepackage{multirow}
\usepackage{textcomp}
\usepackage{caption}
\usepackage{rotating}
\usepackage{sidecap}
\usepackage{ragged2e}
\usepackage{floatrow}
\usepackage{subfig}
\usepackage{dirtytalk}
\sidecaptionvpos{figure}{c}
\usepackage{appendix}

\usepackage{xargs} 

\newcolumntype{C}{>{\centering\arraybackslash}m{0.3\linewidth}}
\newcolumntype{D}{>{\centering\arraybackslash}m{0.13\linewidth}}
\newcolumntype{E}{>{\centering\arraybackslash}m{0.16\linewidth}}
\newcolumntype{f}{>{\centering\arraybackslash}m{0.4\linewidth}}
\newcolumntype{G}{>{\centering\arraybackslash}m{0.20\linewidth}}
\newcolumntype{H}{>{\centering\arraybackslash}m{0.15\linewidth}}
\newcolumntype{I}{>{\centering\arraybackslash}m{0.14\linewidth}}
\newcolumntype{J}{>{\centering\arraybackslash}m{0.17\linewidth}}
\newcolumntype{K}{>{\centering\arraybackslash}m{0.18\linewidth}}
\newcolumntype{M}{>{\centering\arraybackslash}m{0.18\linewidth}}
\newcolumntype{W}{>{\centering\arraybackslash}m{0.08\linewidth}}
\newcolumntype{Z}{>{\justify\arraybackslash}m{0.8\linewidth}}
\newcolumntype{X}{>{\justify\arraybackslash}m{0.18\linewidth}}
\newcommand{\tabitem}{~~\llap{\textbullet}~~}
\crefformat{section}{\S#2#1#3}

\newif\ifcomments
\commentstrue

\author{ Isidora Chara Tourni\\ {\bf Lei Guo} \\{\bf Hengchang Hu}\And \\{\bf Edward Halim}\\{\bf Prakash Ishwar}  \\ \And {\bf Taufiq Daryanto}\thanks{\ \ Institut Teknologi Bandung } \\{\bf Mona Jalal}  \\ \\Boston University\\ \texttt{wijaya@bu.edu} \And \\ {\bf Boqi Chen}\\ {\bf Margrit Betke}\And {\bf Fabian Zhafransyah}\footnotemark[1] \\ {\bf Sha Lai} \\ {\bf Derry Tanti Wijaya}\thanks{\ \ Corresponding Author } \\ }

\makeatletter\let\sf@counterlist\@empty\makeatother
\begin{document}

\title{Detecting Frames in News Headlines and Lead Images in U.S. Gun Violence Coverage}
\maketitle

\begin{abstract}
News media structure their reporting of events or issues using certain perspectives.  
When describing an incident involving gun violence, for example, some journalists may focus on mental health or gun regulation, while others may emphasize the discussion of gun rights. Such perspectives are called \say{frames} in communication research. 
We study, for the first time, the value of combining lead images and their contextual information with text to identify the frame of a given news article.

We observe that using multiple modes of information(article- and image-derived features) improves prediction of news frames over any single mode of information when the images are relevant to the frames of the headlines. We also observe that frame image relevance is related to the ease of conveying frames via images, which we call  \textit{frame concreteness}. 

Additionally, we release the first \textit{multimodal} news framing dataset related to gun violence in the U.S., curated and annotated by communication researchers. The dataset will allow researchers to further examine the use of multiple information modalities for studying media framing.
\end{abstract}

\section{Introduction}

Media framing refers to the journalistic practice of selecting aspects of a perceived reality and making them more salient in news coverage~\cite{Entman93,ReeseGaGr01}. In political communication, for example, news framing is helpful as it reveals how the news article is structured to promote a certain side of the political spectrum, thus influencing the public opinion in a particular way.

Journalists have been using both text and images to frame news stories. Images in news stories can help convey controversial or provocative meanings that would otherwise be unpalatable to the news audience, if it were spelled out in text~\cite{MessarisAb01}. While text is more influential in changing opinions, visuals elicit more attention and emotional reactions, resulting in behavioral change~\cite{ColemanWu15,Dan17,PowellBoDeDe15}. 
Lead images may carry additional background knowledge about the event (e.g., showing well-known people and locations).  An image showing a school, for example, might suggest an article about gun violence focuses on the \say{School/Public Safety} frame.

Text and images thus work in tandem to create a holistic perception of news and must be considered together when analyzing news frames ~\cite{WesslerWoHoLu16}.

Given the importance of visuals in media framing and the rising gun violence in the U.S. \cite{guo2021makes}, we extend the Gun Violence Frame Corpus (GVFC) \cite{SLiu19}, which contains news headlines related to U.S. gun violence and their domain-expert frame annotations, by retrieving the lead images of the articles and obtaining their relevance annotations from communication domain experts 
(i.e., an image is annotated as \textit{relevant} if it expresses the annotated headline frames). Notably, about half of the time, the images presented do not express the annotated headline frames (Table \ref{tab:numsamples}). This might be explained from the journalism research perspective, as reporters and photographers do not necessarily work together seamlessly in the newsroom as they occupy distinct occupational roles and often compete for control over how a story may be packaged and presented as a final product~\cite{Lowrey02}. 

Hence, in addition to communication scholars benefiting from tools that can analyze, on large scale, images and headlines in tandem for frames, newsroom editors would benefit from tools that can identify images that help depict the main thrust of the story’s focus~\cite{Caple10}. Such tools do not yet exist, and our work addresses this need. 

In this work, we comprehensively explore the use of multimodal information from news articles i.e., headlines or summaries, and their lead images i.e., the images, categories of objects in the images, or the background/real-world knowledge contained in them, for predicting frames. 

Our results show that for news articles with relevant images, using only image-derived features or only article-derived features (i.e., headlines and/or extractive summaries) yields less accurate frame predictions than our multiple modalities approach. When considering articles with irrelevant images, the accuracy of the multimodal approach is comparable to that based on only article-derived features. 

We also observe that adding image contextual information using the Google Web Entity Tagger API\footnote{\footnotesize\url{https://cloud.google.com/vision/docs/detecting-web}} or an entity-aware news image caption generation model \citep{tran2020transform} or by asking humans to annotate the central subject of the image in terms of pre-defined categories that cue gun violence frames, such as politician (\textit{politics} frame), legislative buildings (\textit{gun control} frame), school/campus (\textit{school/public safety} frame), etc., improves the performance of frame prediction, compared to using raw images alone. The API tags capture background information associated with an image from the Web, such as the list of named entities in the image, by finding similar images in the Web and parsing the associated web page contents. News image captions capture real-world information in the image, e.g., the names of people and objects, by learning to associate words in the article text with faces and objects. Human annotations of the central subject of the image in terms of categories such school or legislative buildings, capture the annotators' background knowledge of the identities of entities in the image. 

Overall, our contributions are the following:
(1) A well-curated \textit{multimodal text-image framing dataset with expert annotations}\footnote{The multimodal annotation is available for download through the GVFC dataset website: \url{https://derrywijaya.github.io/GVFC.html}.}: With the goal of predicting frames based on multiple information modalities, we augment GVFC by using the article URLs to retrieve lead images of articles and annotate the images for their visual framing labels, which include (a) the Subject/Race/Ethnicity (SRE) annotations of the central \textit{subject} of the image (i.e., suspect vs.~victim vs.~politician, etc.) and whether the image contains anything related to \textit{race/ethnicity}, and (b) the image relevance annotations i.e., whether the images are relevant to the annotated frames of their headlines. 
In addition to frame annotations of news article headlines and lead images, for each image we provide its URL, Web Entity tag (API tag), caption generated using a state-of-the-art news image captioning system \cite{tran2020transform}, and the article summary generated by a state-of-the-art extractive summarization system \cite{lapata-2019-text}. 
(2) \textit{Comprehensive study and development of methods to combine multimodal information to predict article frames and image relevance}: We explore various approaches to predict image relevance and article frames using information from both article and lead images, using BERT  \cite{DevlinChLeTo18} to represent text and a deep convolutional neural
network ResNet-50 \cite{HeZhReSu16} to represent raw images.
(3) \textit{Frame concreteness}: We propose a novel method for measuring the ease of conveying frames through images via the concreteness of words in its headlines, i.e., the ease of identifying tangible concepts and mental images that arise in correspondence to words \cite{paivio1968concreteness}, and relate frame image relevance to frame concreteness. 

\setlength{\textfloatsep}{5pt}
\begin{figure}[t]
\begin{center}
\includegraphics[width=0.80\textwidth]{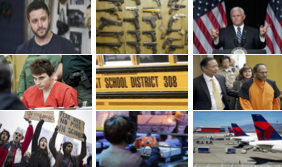}
\end{center}
\caption{\small{Sample images for each frame (from left to right and top to down): 2nd Amendment, Gun Control, Politics, Mental Health, School/Public Space Safety, Race/Ethnicity, Public Opinion, Society/Culture, Economic Consequences.}}
\label{fig:9leadImages}
\end{figure}

\section{Related Work}
Media framing is related to many factors such as word choice, the presentation of background information, and the emphasis on certain actors.

The subtle nature of news framing can influence the opinion of readers in a certain way without them even noticing it, hence its analysis has many implications, e.g., it has been used to understand why important public affairs issues such as gun violence are polarizing \cite{SLiu19}, how media manipulation strategies are conducted \cite{FieldKlWiPaJuTs18}, or how framing is used to perpetuate racial biases \cite{drakulich2015explicit}.  In Journalism, most ``framing analyses'' have been done manually
hence they are not scalable \cite{HamborgZhGi19}.

In natural language processing (NLP), automated frame detection focuses on predicting frames from news texts. Some of these methods rely on topic models \cite{NguyenBoRe13}.
\citet{NaderiHi17} devised various deep neural networks (LSTMs, BiLSTMs, or GRUs) for predicting frames at the sentence level in the  Media Frame Corpus (MFC)~\cite{CardBoGrReSm15}.  The most recent method for predicting news frames in headlines is the work of \citet{SLiu19,akyurek-etal-2020-multi}. They fine-tuned BERT \cite{DevlinChLeTo18} using focal loss \cite{lin2017focal} to predict frames of headlines.
They also released a framing benchmark dataset, the Gun Violence Frame Corpus (GVFC), of news links and headlines with frame annotations, related to gun violence in the U.S. 
They show that fine-tuning BERT for frame prediction using news headlines results to significantly higher accuracy than previous methods.

In this paper, we use GVFC articles' lead images and perform experiments with a rich set of unimodal and multimodal information to predict frames. Although images and text have been used together as multimodal inputs to improve performance in other NLP tasks such as machine translation \cite{specia2016shared,hewitt2018learning,caglayan2019probing,yao2020multimodal,khani2021cultural} or in vision-language tasks such as multilingual image retrieval or captioning \cite{kim2020mule,burns2020learning,rasooli2021wikily}, the use of images and text in tandem to \textit{automatically} analyze framing i.e., computational multimodal framing has never been explored--all previous works in multimodal framing have been conducted manually \cite{MessarisAb01,ColemanWu15,Dan17,PowellBoDeDe15,WesslerWoHoLu16}. Given the growing importance of visual journalism and the contribution of images to media framing which suggest that images may be able to help interpret text, our work is the first to conduct computational multimodal framing analysis, which will enable scalable multimodal framing analysis. 

\section{Dataset}
\label{sec:dataset}
\vspace{-2mm}
Our multimodal version of GVFC contains news headlines and their corresponding lead images, news URLs, and the entire news text.

The lead images are either the pictures shown at the top of news articles or the editor-picked thumbnails that are shown in news services such as Google News (Fig.~\ref{fig:9leadImages}).
Using Brandwatch Consumer Research\footnote{\footnotesize\url{https://www.brandwatch.com}} we analyze 3,000 news headlines, of which 1,300
are annotated with 9 major frames, e.g., politics, gun control/regulation, mental health, race/ethnicity, etc (Table~\ref{tab:numsamples}), that exhaustively cover the discussion of the U.S. ``gun violence'' issue in communication research. In this paper, we further annotate each lead image with a binary relevance label that indicates whether the image is consistent with the frame associated to the headline. 

We also annotate the central subject (S) of the image using one of 16 categories that often imply certain frames, e.g., suspect (often implies mental health), politician (often implies politics), company logos (often implies economic consequence), etc. (see Appendix for the full listing), and an additional race/ethnicity (RE) label which can take one of the following 3 values: 
1) racial/ethnic minority groups, 2) hate groups, or 3) none of the above. 

The details of the visual annotation codebook used to train the coders are given in the Appendix. The coders' agreement on how to apply the codes is
measured with inter-coder reliability (ICR). High ICR values (above
90\% agreement or 0.70 Krippendorff $\alpha$ \cite{krippendorff2018content}) imply that two or more coders consistently categorized the content similarly, signaling a high validity of the results. In our dataset, ICR was met on all variables: Subject (90\% agreement, 0.88 $\alpha$),
Race/Ethnicity (91\% agreement, 0.64 $\alpha$) and Relevance
(88\% agreement, 0.75 $\alpha$). The number and ratio of relevant images per frame are shown in Table~\ref{tab:numsamples}.
Easy and hard to classify examples and their features are provided in Table~\ref{table:fig_examples} and described in detail in \cref{sec:experiments}.
We affirm we have the right to use the collected dataset in the way we are using it\footnote{We have confirmed and received approval from Brandwatch Consumer Research whom we obtain the data from.}, i.e. the article headline and URL, as well as their annotations and image-derived visual and textual features; and we bear responsibility in case of a violation of rights or terms of service. Researchers can use the article URLs to retrieve images and full texts of the articles. 

\setlength{\textfloatsep}{5pt}
\begin{table}[t]
\centering
\scriptsize
\begin{tabular}{lcc} 
\toprule
 &  & \textbf{\# Relevant} \\
\textbf{News Frame} & \textbf{\# Articles} & \textbf{Images (\%)} \\
\midrule
Politics & 373 & 241 (65\%)\\
Public Opinion & 237 & 147 (62\%) \\
Gun Control/Regulation & 215 & 93 (43\%)\\
School/Public Space Safety & 137 & 68 (50\%)\\
Economic Consequences & 80 & 46 (58\%)\\
Race/Ethnicity & 114 & 34 (30\%) \\
Mental Health & 65 & 28 (43\%)\\
2nd Amendment/Gun Rights & 38 & 13 (34\%)\\ 
Society/Culture & 41 & 4 (10\%) \\
\midrule
Overall & 1,300 & 674 (52\%)\\
\bottomrule
\end{tabular}
\caption{\small {Gun violence frames in our dataset, the number of articles with headlines and lead images, and the  number of lead images annotated as relevant to the frame with the percentage indicated in brackets. The news frames are ordered by the number of relevant images from highest to lowest. 
}
}
\label{tab:numsamples} 
\end{table}

\begin{table*}[ht!]
\setlength{\tabcolsep}{3pt} 
\renewcommand{\arraystretch}{0.5} 
\begin{tabular}{X Z}
\toprule
\textbf{\small{Image}} & \textbf{\small{Description}}
\\
\midrule
{\vspace{-8mm}
\centering
\includegraphics[height=0.111\textwidth,width=0.18\textwidth]{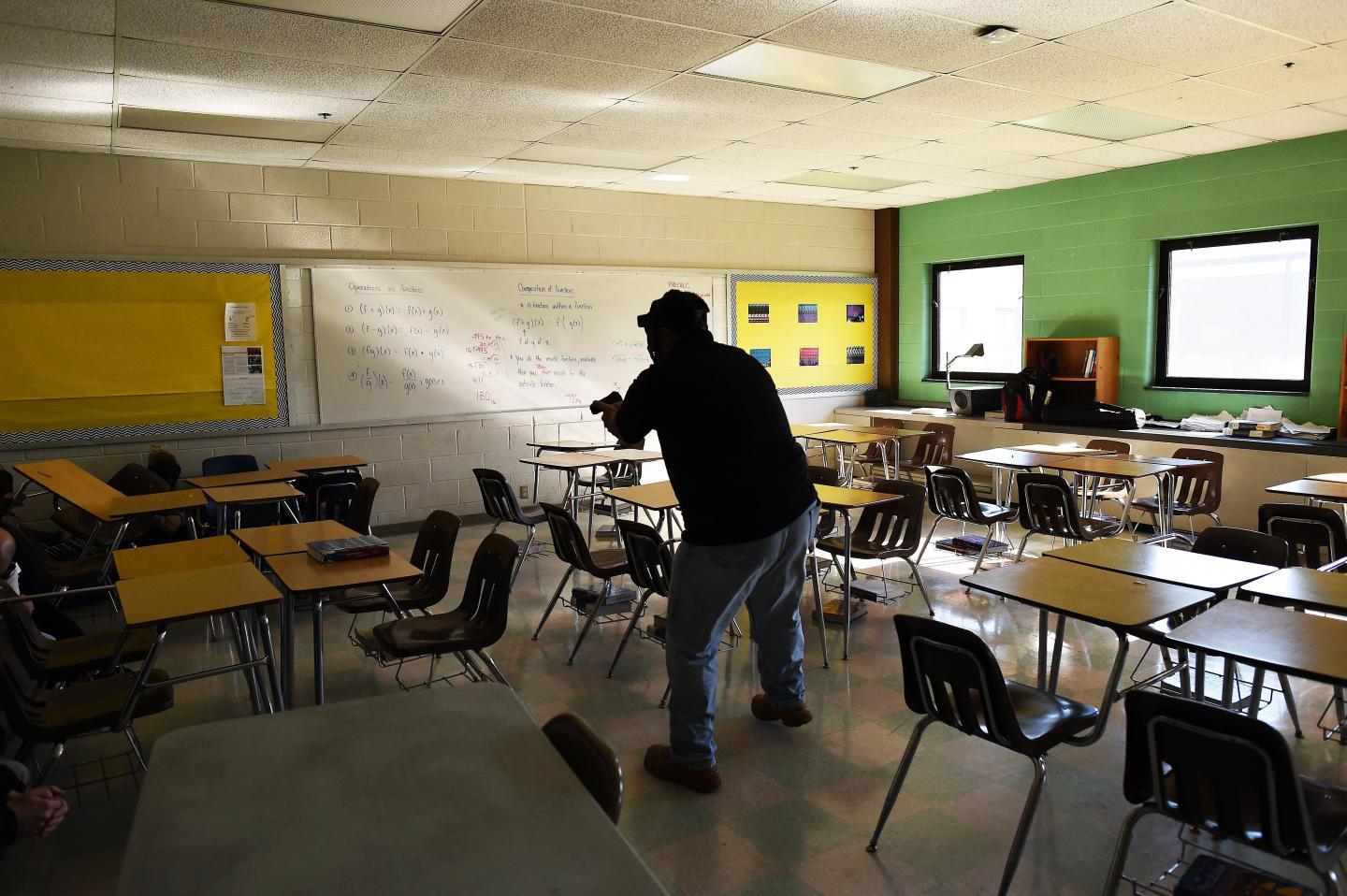}} & \vspace{-1mm}{\scriptsize\underline{An article with \textbf{relevant} image in a \textbf{frame with many examples} (potentially \textbf{easy} to classify)}\vspace{-1mm}\newline \tiny{\textbf{\textit{\scriptsize{Frame}}}: School/Public Space Safety \vspace{-2mm}\newline {\tiny\textbf{{\textit{\scriptsize{Headline}}}}: To Defend Against School Shootings, Massachusetts District Is Passing Out Emergency Buckets With Hammer, Rope} \vspace{-2mm}\newline  {\tiny{\textbf{\textit{\scriptsize{API tag}}}: Classroom, School, Harry S Truman, High School, Active shooter, Lockdown, Campus, Student}} \vspace{-2mm}\newline {\tiny{\textbf{\textit{\scriptsize{Caption}}}}: A school shooting victim in Brockton, Mass., last month.} 
\vspace{-2mm}
\newline{\tiny{\textbf{\textit{\scriptsize{3sentences}}}}: \vspace{-2mm}U.S.: More than 1,000 blue buckets were assembled to be passed out to classrooms in the Brockton, Massachusetts, school district, filled \vspace{-2mm}with curated items aimed at saving lives in the event of an emergency, including a school shooting.
The Brockton school district partnered with the 
mayor’s office, the Brockton Police Department and a local Lowe’s to put together buckets filled with four items to help defend classrooms. Each blue five-gallon bucket contains a wooden wedge, a one-pound hammer, a 50-foot length of rope and a roll of duct tape, according to The Enterprise.}
\vspace{-2mm}\newline {\tiny{\textbf{\textit{\scriptsize{Summary}}}}: \vspace{-2mm}U.S.: More than 1,000 blue buckets were assembled to be passed out to classrooms in the Brockton, Massachusetts, school district. The buckets can be used for emergency bathroom situations. Mayor Bill Carpenter applauded the decision to put the buckets in the classrooms.}
}} 
\\
\midrule
{\vspace{-2mm}
\raggedright
\includegraphics[height=0.111\textwidth,width=0.18\textwidth]{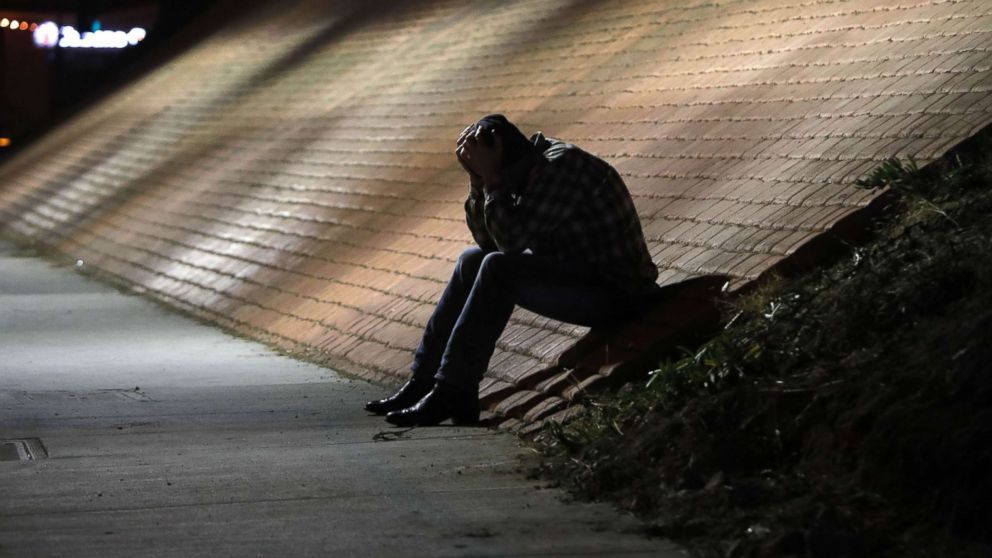}}& \vspace{-1mm}{\scriptsize\underline{An article with \textbf{irrelevant} image in a \textbf{frame with many examples} (potentially \textbf{harder} to classify).}\vspace{-1mm}\newline \tiny{\textbf{\textit{\scriptsize{Frame:}}} School/Public Space Safety \vspace{-2mm}\newline \textbf{\textit{\scriptsize{Headline}}}: Mass shootings `increasing' and pose `most serious threat' in US, expert says, \vspace{-2mm}\newline \textbf{\textit{\scriptsize{API tag:}}} Thousand Oaks shooting, Borderline Bar Grill, Mass shooting, California Bar, Police officer \vspace{-2mm}\newline \textbf{\textit{\scriptsize{Caption}}}: A gunman at the scene of the shooting at a country bar in Sacramento.
\vspace{-2mm}\newline {\tiny{\textbf{\textit{\scriptsize{3sentences}}}}: \vspace{-2mm}Mass shootings `increasing' and pose `most serious threat' in US, expert says At least 59 people have been killed as a result of mass shootings this year. Deadliest mass shootings of 2018 in the U.S. Mike Nelson/EPA via Shutterstock.}
\vspace{-2mm}\newline {\tiny{\textbf{\textit{\scriptsize{Summary}}}}: \vspace{-2mm}At least 59 people have been killed as a result of mass shootings this year. There have been at least six mass shootings in the U.S. this year, according to the U.S., at least 10 mass shootings have been linked to mass shootings at a California bar.}
}}
\\
\midrule
{\vspace{-13mm}\raggedright \includegraphics[height=0.111\textwidth,width=0.18\textwidth]{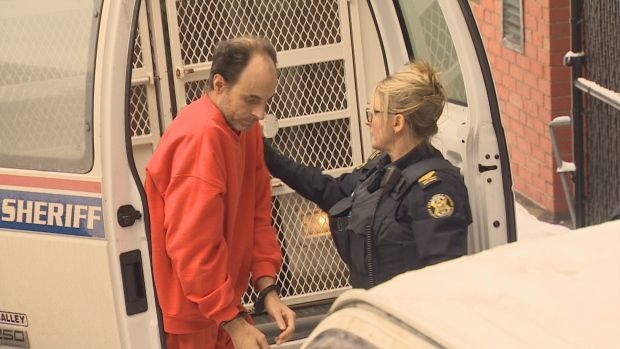}}& \vspace{-1mm}{\scriptsize\underline{An article with \textbf{relevant} image in a \textbf{frame with few examples} (potentially \textbf{harder} to classify).}\vspace{-1mm}\newline \tiny{\textbf{\textit{\scriptsize{Frame:}}} Mental Health \vspace{-2mm}\newline \textbf{\textit{\scriptsize{Headline}}}: Accused Fredericton shooter will undergo psych assessment \vspace{-2mm}\newline \textbf{\textit{\scriptsize{API tag:}}} Car, Job, Vehicle, Staff, Capilar y Corporal \vspace{-2mm}\newline \textbf{\textit{\scriptsize{Caption}}}: Matthew Vincent Raymond Murder Officer Suspect Broward Police officer Arrest \vspace{-2mm}warrant Criminal charge Suspect Murder.
\newline {\tiny{\textbf{\textit{\scriptsize{3sentences}}}}: \vspace{-2mm}The Fredericton man accused of killing four people in August will be sent for a psychiatric assessment. Judge Julian Dickson ordered the assessment Wednesday to determine if Matthew Vincent Raymond, 48, \vspace{-2mm}is fit to stand trial on four counts of first-degree murder. Raymond is charged in the Aug. 10 shooting deaths of Fredericton police constables Robb Costello, 45, and Sara Burns, 43, and civilians Donnie Robichaud, 42, \vspace{-2mm}and Bobbi Lee Wright, 32.}
\vspace{-2mm}\newline {\tiny{\textbf{\textit{\scriptsize{Summary}}}}: \vspace{-2mm}A judge orders the assessment to determine if Matthew Vincent Raymond, 48, is fit to stand trial on four counts of first-degree murder. Raymond is charged in the Aug. 10 shooting deaths of Fredericton police constables Robb Costello, 45, and Sara Burns, 43, and civilians Donnie\vspace{-2mm} Robichaud, 42, and Bobbi Lee Wright, 32. Arguments about who should conduct the assessment. \vspace{-2mm}The assessment is expected to be completed before Dec. 4, when Raymond is due back in court.}
}}
\\
\midrule
{\vspace{-15mm}\raggedright \includegraphics[height=0.12\textwidth,width=0.18\textwidth]{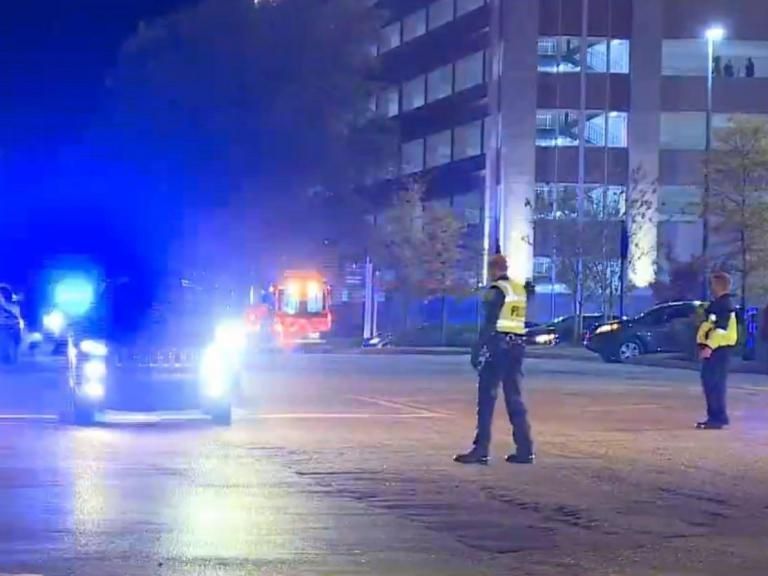}} & \vspace{-1mm}{\scriptsize\underline{An article with \textbf{irrelevant} image in a \textbf{frame with few examples} (potentially \textbf{hardest} to classify).}\vspace{-0.5mm}\newline \tiny{\textbf{\textit{\scriptsize{Frame}}}: Race/Ethnicity \vspace{-2mm}\newline \textbf{\textit{\scriptsize{Headline}}}: Alabama mall shooting: Family of black man killed by police officer on Thanksgiving hires civil rights lawyer \vspace{-2mm}\newline \textbf{\textit{\scriptsize{API tag}}}: \vspace{-2mm}Shooting of Emantic Fitzgerald Bradford Jr., Alabama Shooting of Michael Brown, News, Shopping Centre, Breaking news, Street light, Television show, Street Broken Horses \vspace{-2mm}\newline \textbf{\textit{\scriptsize{Caption}}}: \vspace{-2mm}A police officer at the scene of the shooting at the Riverchase Galleria in Birmingham, Ala., on Friday.
\newline {\tiny{\textbf{\textit{\scriptsize{3sentences}}}}: \vspace{-2mm}Emantic Fitzgerald Bradford Jr ’s family has employed Benjamin Crump who previously represented the families of shooting victims Trayvon Martin and \vspace{-2mm}Michael Brown to also represent them: WVTM The family of a 21-year-old black man who was shot by a police officer at shopping centre in Alabama on Thanksgiving has hired a national civil rights lawyer to represent them.}
\vspace{-2mm}\newline {\tiny{\textbf{\textit{\scriptsize{Summary}}}}: \vspace{-2mm}Emantic Fitzgerald Bradford Jr was fatally shot by a police officer at the shopping centre in Alabama on thanksgiving has hired a national civil rights lawyer. \vspace{-2mm}Police initially said a hoover police officer who was responding to reports of gunfire at a shopping mall confronted an armed man running away from the scene and fatally shot him. \vspace{-2mm}The shots responsible for injuring an 18-year-old man and a 12-year-old girl, but investigators have since said they believe he did not firesle the gunman is still at large. \vspace{-2mm}Mr Crump said Bradford was a veteran who was licensed to carry a concealed firearm. Bradford’s family said they are working with our legal team to determine.}
}}
\\
\bottomrule
\end{tabular}
\caption{\small{Examples of articles, their images and image- and article-derived textual features that are potentially easy or hard to classify using the multimodal approach.}}
\label{table:fig_examples}
\end{table*}

\section{Experiments}
\label{sec:experiments}
We experiment with unimodal (\cref{sec:unimodal_text}) and multimodal (\cref{sec:multimodal}) information obtained from each article and its lead image. We train and report 4-fold cross validation frame prediction accuracies (Table~\ref{table:classification_acc}) for all articles in our dataset (\textit{All Articles}), and for the subset of articles with relevant images (\textit{Articles with Relevant Images}). 
We also perform image-to-frame relevance classification (\cref{sec:relevance}). 

\subsection{Unimodal}
\label{sec:unimodal_text}
In this set of experiments we predict news frames from only one mode of information, either image-derived or article-derived. 
For raw images, we use ResNet-50 \cite{HeZhReSu16} for both image representation and frame classification (\textsc{ResNet-50}). We use BERT \cite{vaswani2017attention} to represent text from the article headlines (\textsc{BERT headline}), the image Web Entity API tags 
(\textsc{BERT API}), or the automatically generated captions from images 
(\textsc{BERT Caption}).
We also experiment with other article-derived information in the form of text: the headline concatenated with the automatically generated extractive summary (\textsc{BERT headline + Summary}), or with the first three sentences of the article (\textsc{BERT headline + 3sentences}), a typical baseline for extractive summarization.
For all text-form information derived from the article (irrespective of whether the text was extracted from the image, headline or body of the article), we follow the state-of-the-art methodology for frame detection based on news headlines \cite{SLiu19}, which constitutes our baseline (\textsc{BERT headline}). Specifically, we use the text as input into BERT's pre-trained base uncased model and fine-tune the model to predict the frames of the articles over 25 different random seeds to avoid the fine-tuning instability due to the small dataset size \citep{devlin-etal-2019-bert, dodge2020fine, mosbach2021on}. In all our models, the number of epochs is 10, the batch size is 4, and the learning rate is 2e-5.

\noindent{\bf Using Image-derived \textit{Visual} Features} (\textsc{ResNet-50}). We predict news frames based only on the raw lead images of the news articles. We use the ResNet-50 model~\cite{HeZhReSu16}, pre-trained on ImageNet~\cite{DengDoSoLiLiLi09}, and replace the output layer of the original ResNet-50 network with a flattened layer of 512 nodes followed by a dropout layer with a 0.5 dropout rate to the frame classification (9-nodes) layer.

All images are scaled to $224\times224$ pixels and are normalized based on the mean and standard deviation of ImageNet.

\noindent{\bf Using Image-derived \textit{Visual} Annotations} (\textsc{SRE}). We create a 19-length feature vector for each image obtained from the Subject, Race/Ethnicity (SRE) annotations of the image that indicate the human coders' background knowledge of the image's central Subject, and its connection to Race/Ethnicity. We train a logistic regression frame classifier with this feature vector as input. 

\noindent{\bf Extracting Image-derived \textit{Textual} Features: Google Web Entity API Tags } (\textsc{BERT API)}. Here, frames are predicted based only on the Google Web Entity tags of lead images. Web Entity detection is a Google cloud service that reads an image as input and returns a ranked list of web entities as tags. For each image, we form a \say{sentence} by concatenating the top-10 Web Entity tags returned for the image.

\noindent{\bf Extracting Image-derived \textit{Textual} Features: Image Captions} (\textsc{BERT Caption}).

We follow \citet{tran2020transform} to generate captions for the lead images of news articles. The model introduced in the paper consists of different encoders generating representations for each modality (article text, images, faces, and objects), and a Transformer as the decoder attending over text, images, image faces and objects. 

It uses Byte-Pair-Encoding, breaking sequences into subwords and then merging common sequences into larger words. This leads to better generalization and prediction of out-of-vocabulary words and names, and ultimately to linguistically rich captions for images accompanying each news article. 

We follow all default settings and parameters suggested in the paper, and use RoBERTa \cite{liu2019roberta} as the article text encoder, a ResNet-152 \cite{dauphin2017language} pretrained on ImageNet as the image encoder, MTCNN \cite{zhang2016joint} as the face detector, and YOLOv3 \cite{farhadi2018yolov3} as the object detector, with the latter two operating as the specialized image face and object attention modules, respectively. All representations obtained from the individual encoders are fed into a four block Transformer decoder, which employs a multi-head multi-modal attention mechanism and generates byte-pair encoded tokens, that are finally concatenated to form the caption.

\noindent{\bf Extracting Article-derived \textit{Textual} Features: Summary} (\textsc{BERT Summary}) 

We automatically extract the summary of the article, following \citet{lapata-2019-text}, that uses BERT to represent sentences, and inter-sentence Transformer layers on top of the BERT encoder to classify whether a sentence should be in the extractive summary.  

\subsection{Multimodal}
\label{sec:multimodal}
In this set of experiments, we predict news frames using multiple modes of information derived from both the  image and the article. 
\noindent{\bf Using Image-derived \textit{Visual} and \textit{Textual} Features and Article-derived \textit{Textual} Features} (\textsc{ResNet-50 + BERT Headline}, \textsc{ResNet-50 + BERT Headline + API}, \textsc{ResNet-50 + BERT Headline + Caption}). 
Here, frames are predicted with multiple input modalities (visual \textit{and} textual features). 
We follow a simple concat fusion approach, which allows us to build a modular pipeline, obtain the text and visual representations from their respective modules, and use them to predict the frame class. Specifically, we use \textsc{ResNet-50} representations of the raw images and, as suggested by \citet{devlin-etal-2019-bert} for best performance, representations of the text obtained from the concatenation of the contextual embeddings of the last four layers of BERT, which has been fine-tuned for frame classification, as inputs to our multimodal 
3-layer feed forward fully connected classifier neural network that we \textit{then} train jointly with \textsc{ResNet-50}. 

We use the AdamW optimizer,
cross entropy loss, 
and ``no improvement in validation accuracy over 5 epochs'' as the stopping criterion.

\noindent{\bf Using Image-derived \textit{Visual} Annotations and Article-derived \textit{Textual} Features} (\textsc{BERT Headline + SRE}). We concatenate BERT representation of the headline with the SRE feature vector and train jointly with fine-tuning BERT using a 1-layer feed forward fully connected classifier neural network added on top of BERT.

\noindent{\bf Using Image-derived \textit{Textual} Features and Article-derived \textit{Textual} Features} (\textsc{BERT Headline + API}, \textsc{BERT Headline + Caption}). Here we concatenate article headlines and image-derived textual features (API tags or captions) as input to fine-tune BERT for frame classification.

\vspace{-1mm}
\subsection{Relevance}
\label{sec:relevance}
We use the article headline (\textsc{BERT Headline}) and the image-derived features (\textsc{BERT API}, \textsc{BERT Caption}), the SRE annotations (\textsc{SRE}), and their combinations (\textsc{BERT Headline + API}, \textsc{BERT Headline + Caption}, \textsc{ResNet-50 + BERT Headline}, \textsc{ResNet-50 + BERT Headline + API}, \textsc{ResNet-50 + BERT Headline + Caption}) to predict the relevance of the images to the frames of their headlines.
We perform a 4-fold cross-validation binary classification for relevance prediction, with the same BERT and ResNet-50 architectures and hyperparameters as before. Accuracies 
with uni- and multimodal information sources are reported in Table~\ref{table:relevance}.
To mimic the relevance annotation process of our coders, who are given the labeled frame of the headline to decide whether the lead image is relevant to it, we provide our relevance prediction model with headline frame. We concatenate the frame label to the input of the top performing models of Table~\ref{table:relevance} and their combinations, and report accuracies in Table \ref{table:relevance_frame}. 

\begin{table*}
\centering
\Large
\resizebox{\textwidth}{!}{
\begin{tabular}{cDDDDDDDDDDDDDD}
\toprule
\textbf{\Large{Method}}&\textbf{\Large{ResNet-50}}&\textbf{\Large{SRE}}&\textbf{\Large{BERT headline}} & \textbf{\Large{BERT API}}& \textbf{\Large{BERT Caption}}  &\textbf{\Large{BERT headline + API}}&\textbf{\Large{BERT headline + Caption}}&\textbf{\Large{BERT headline + Summary}}&\textbf{\Large{BERT headline + 3sentences}}&\textbf{\Large{ResNet-50 + BERT headline}} &\textbf{\Large{ResNet-50 + BERT headline +  API}}&\textbf{\Large{ResNet-50 + BERT headline + Caption}}&\textbf{\Large{BERT headline + SRE}}\\
\midrule
\multicolumn{13}{l}{{\textbf{{All Articles}}}} \\ 
\midrule
\Large{}&\Large{9.3}&\Large{49.2}&\Large{81.9}&\Large{47}&\Large{48.1}&\Large{82}&\Large{82}&\Large{82.4}&\Large{81.8}&\Large{13.8}&\Large{13.7}&\Large{12.2}&\Large{81.5}\\
\midrule
\multicolumn{13}{l}{{\textbf{{Articles with Relevant Images}}}} \\ 
\midrule
\Large{}&\Large{42.8}&\Large{81.2}&\Large{83}&\Large{72.1}&\Large{72.5}&\Large{87}&\Large{84.6}&\Large{83.1}&\Large{83.1}&\Large{49.7}&\Large{65.3}&\Large{63.8}&\Large{83.2}\\
\bottomrule
\end{tabular}}
\caption{\small{Overall micro accuracy of our methods for \textbf{frame classification} for \textit{All articles} and \textit{Articles with Relevant Images}. \textsc{BERT headline} is the baseline we compare to \cite{SLiu19}}}
\label{table:classification_acc}
\end{table*}

\begin{table*}
\centering
\resizebox{\textwidth}{!}{
\begin{tabular}{ccccccccccc}
\toprule
\textbf{\huge{}} & \textbf{\huge{BERT}}&\textbf{\huge{}}& \textbf{\huge{BERT}} &\textbf{\huge{BERT}} &\textbf{\huge{BERT headline}}&\textbf{\huge{BERT Headline}}&\textbf{\huge{ResNet-50}}&\textbf{\huge{ResNet-50}}&\textbf{\huge{ResNet-50 }}\\
\textbf{\huge{Method}}  & \textbf{\huge{headline}}&\textbf{\huge{SRE}}& \textbf{\huge{API}} &\textbf{\huge{Caption}} &\textbf{\huge{+ API}}&\textbf{\huge{+ Caption}}&\textbf{\huge{+ BERT headline}}&\textbf{\huge{+ BERT headline + Caption}}&\textbf{\huge{+ BERT headline + API}}\\
\midrule
\huge &\huge{62}
&\huge{68.1}
&\huge{59.3}&\huge{62.5}&\huge{65.8}
&\huge{65.7}
&{\huge{55.5}}&\huge{60}&\huge{58.3}\\
\bottomrule
\end{tabular}
}
\caption{\small Overall micro accuracy of our methods for image \textbf{relevance classification} (without frame label) for \textit{All articles}. }
\label{table:relevance}
\end{table*}

\begin{table}
\renewcommand{\arraystretch}{0.3} 
\centering
\resizebox{\textwidth}{!}{
\begin{tabular}{cccc}
\toprule
\textbf{\small{}} &
\textbf{\small{BERT headline}}&\textbf{\small{}}&\textbf{\small{SRE}}\\
\textbf{\small{Method:}}&
\textbf{\small{+ API tag}}&\textbf{\small{SRE}}&\textbf{\small{+ BERT headline}}\\
\textbf{\small{}} &
\textbf{\small{+ Frame}}&\textbf{\small{+ Frame}}&\textbf{\small{+ API tag + Frame}}\\
\midrule\\
\small &
\small{74.2}&
\small{71.0}&
\small{72.0}\\
\bottomrule
\end{tabular}
}
\caption{\small Overall micro accuracy of our methods for image \textbf{relevance classification with frame} for \textit{All articles}.}
\label{table:relevance_frame}
\end{table}

\section{Discussion of Results}
Despite the challenges of a highly nuanced multi-class frame identification and an intrinsically imbalanced dataset, we achieve a high prediction  accuracy of up to 87\% for \textit{Articles with Relevant Images}, and 82.4\% for \textit{All Articles} (Table~\ref{table:classification_acc}).
It is instructive to examine the utility of article- and image-derived features, and a fusion of all in cases where the lead image is relevant to the article headline. We observe that contextual information derived from the image, in the form of API tags or a caption, along with the article headline, can drive the article perspective more clearly. 

The headline + API tags combination provides the best performance (87\%), compared to image-only 
(43\%),
the SRE image annotation (
81.2\%
), or the the article headline (
83\%) which is a strong unimodal baseline. Even when considering examples with irrelevant images, adding API to headlines does not hurt performance and is comparable to using headlines alone, which is unlike SRE whose performance drops significantly for articles with irrelevant images as these annotations are designed with relevant, i.e., frame-implying images in mind. Furthermore, SRE requires training another model to produce these annotations automatically, which is not trivial as they capture real-world knowledge of the subjects in the image, e.g., whether the person in the image is a politician (cuing \textit{politics} frame) or a gun activist/NRA representative (cuing \textit{2nd Amendment}). These findings, namely that API tags, captions, or SRE yield higher accuracy than the raw image alone, indicate the importance of the contextual or background knowledge of the lead image in driving the news frame
This strongly suggests that the highly nuanced task of frame prediction is challenging using 
images in isolation. 
Our findings also confirm previous observations that training with multiple input modalities, e.g., both visual and textual inputs is hard as each modality may generalize differently and hence underperform when trained jointly~\cite{wang2020makes}.

In terms of relevance prediction, the performance is highest for models supplied with frame labels, mimicking the relevance annotation process. Given a headline and a frame, our method can correctly predict the relevance of an image to the frame with 74\% accuracy using the image's API tag. 
Without frame labels, however, the accuracy of the top-performing method drops to 68.1\% and is based on SRE only. While several SRE categories are strong indicators for certain frames, e.g., the presence of demonstrators suggests the  \textit{public opinion} frame, the use of SRE at inference time necessitates the training of another model for predicting these annotations, which is not trivial. 

Examining the content of the API tags and captions,
we observe that API tags have significantly more proper nouns (71\% to 29\% of all words in tags/captions) and named entities (53\% to 31\%) than captions. On the other hand, captions have more common nouns and verbs. 
Since the performance of frame prediction for articles with relevant images is higher when using headline and API compared to headline and caption, this suggests that frames can be directly cued by lexical items such as proper nouns or named entities, e.g., politicians' names cue politics frame \cite{mendelsohn2021modeling}. Since models may lack real-world knowledge required to identify these, especially when there is insufficient text evidence, e.g., when using headlines, API tags that provide this background knowledge from images facilitate frame prediction. 

\begin{figure}[ht!]
  \centering
\includegraphics[scale=0.38,trim=1mm 1mm 1mm 1mm ,clip]{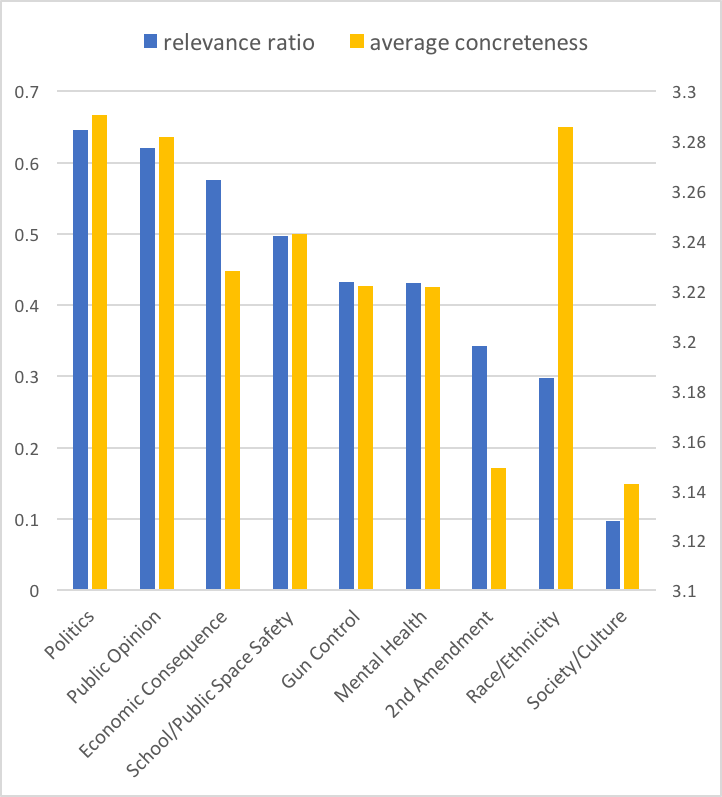}
  \vspace{-4mm}
  \caption{\small Frame relevance ratio and average concreteness.}
\label{fig:concreteness}
\end{figure}
\begin{figure*}
\centering
\begin{subfigure}{.5\textwidth}
  \centering
  \includegraphics[scale=0.14]{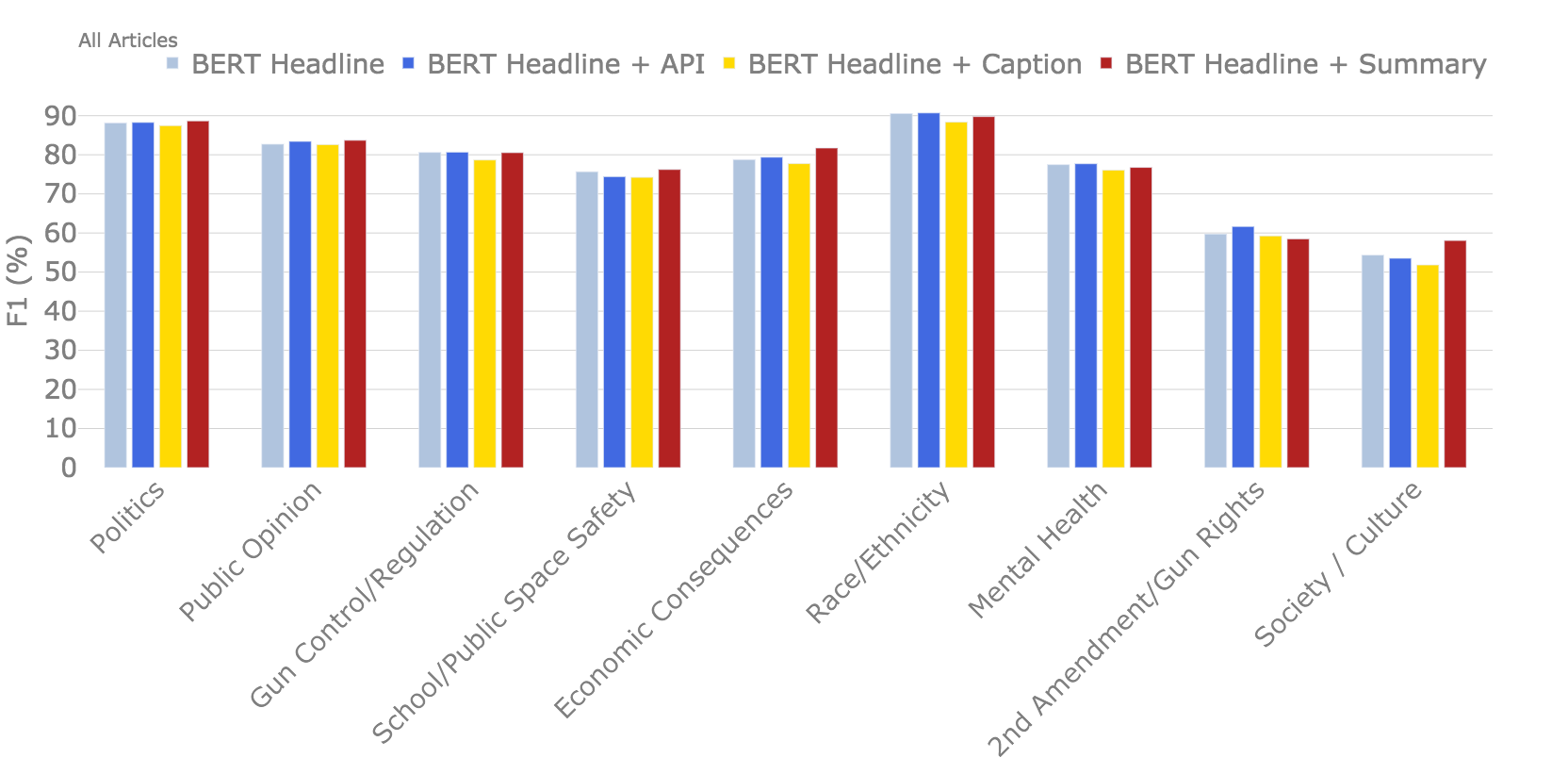}
  \label{fig:all_per_frame}
  \end{subfigure}%
\begin{subfigure}{.5\textwidth}
  \centering
  \includegraphics[scale=0.14]{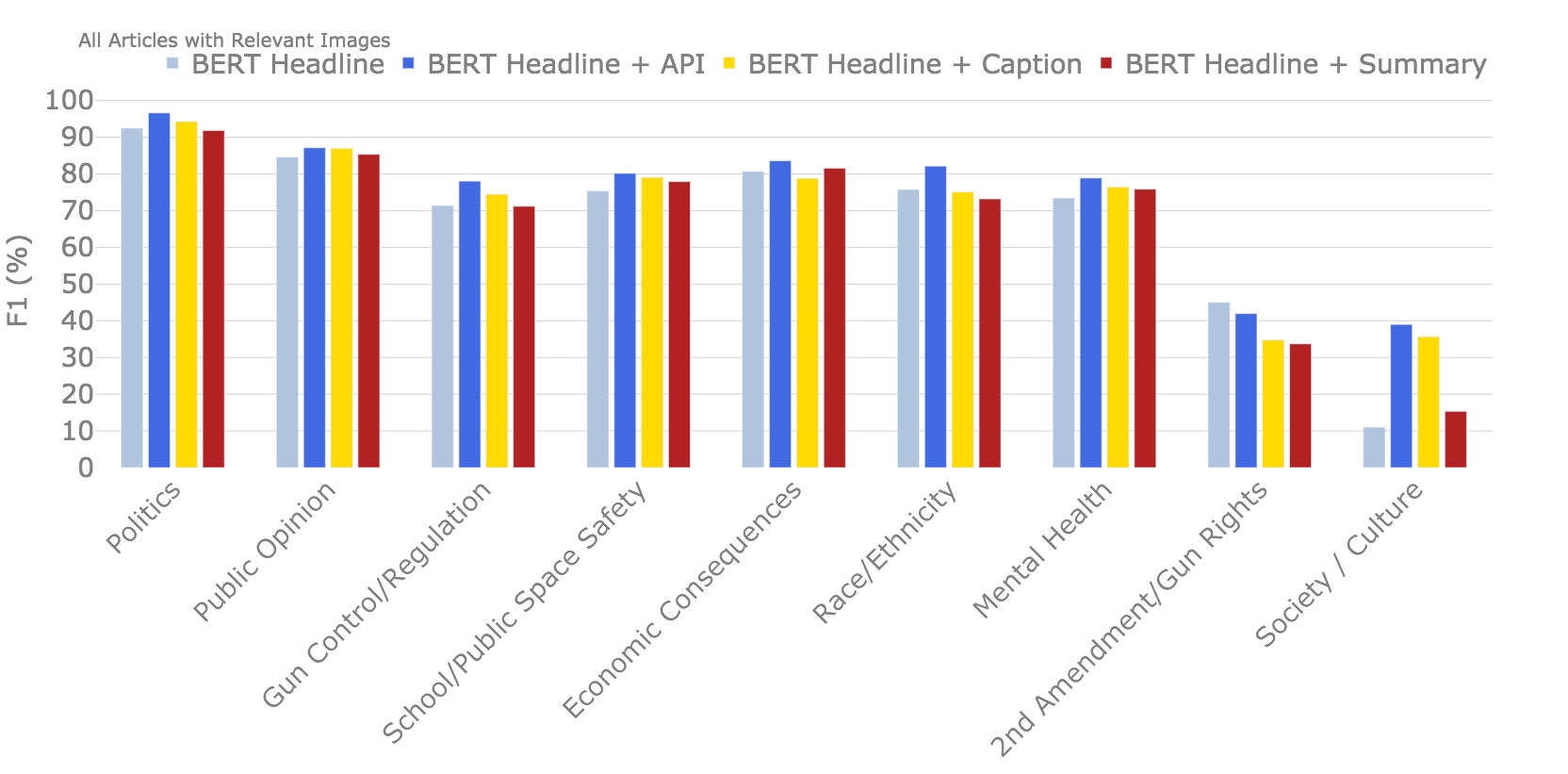}
  \label{fig:rev_per_frame}
\end{subfigure}
\vspace{-9mm}
\caption{\small Per Frame F1 score of the best performing frame prediction methods for (a) \textit{All Articles} (left), (b) \textit{Articles with Relevant Images} (right).}
\label{fig:per_frame}
\end{figure*}
\begin{table*}[ht!]
\setlength{\tabcolsep}{3pt}
\renewcommand{\arraystretch}{0.5} 
\begin{tabular}{p{0.15\textwidth} p{0.2\textwidth} p{0.65\textwidth}}
\toprule
\textbf{\tiny{Error Type}} & \textbf{\tiny{Description}}& \textbf{\tiny{Examples}}\\
\midrule
\tiny{Plausible Interpetation} & \tiny{Predicted frames can be appropriate labels.}&
\tiny{\textit{\underline{\textbf{Example Frame:} Mental Health}}

\textit{\underline{\textbf{Headline:}}} Florida shooter a troubled loner with \textbf{white} supremacist ties

\tabitem{Model erroneously predicted \say{Race/Ethnicity}: \textbf{\say{white}}}

\textit{\underline{\textbf{+ API tag:}}} Nikolas Cruz Stoneman Douglas High School shooting Marjory Stoneman Douglas Murder Mass shooting AR-15 style rifle Suspect Student

\tabitem{Model correctly predicted \say{Mental Health}: \textbf{\say{Suspect Student}}}}\\
\midrule
\tiny{Inferring frames not explicitly cued in text} & \tiny{Predicted frames capture an author's intention without sufficient text evidence.}&
\tiny{\textit{\underline{\textbf{Example Frame:} 2nd Amendment/Gun Rights}}

\textit{\underline{\textbf{Headline:}}} The NRA Versus the Constitution

\tabitem{Model erroneously predicted \say{Gun Control}: \textbf{\say{NRA}}}

\textit{\underline{\textbf{+ API tag:}}} Pennsylvania Obergefell v. Hodges Supreme Court of the United States Concealed carry Rights Reciprocity Act of 2017 Constitution of the United States

\tabitem{Model correctly predicted \say{2nd Amendment}: \textbf{\say{Rights}},\textbf{\say{Constitution}}}}\\
\midrule
\tiny{Missing necessary contextual knowledge} & \tiny{Frames can be directly cued by lexical items (e.g. politicians' names cue Politics frame), yet the model lacks real-world knowledge required to identify those.}&
\tiny{\textit{\underline{\textbf{Example Frame:} Politics}}

\textit{\underline{\textbf{Headline:}}} In closed-door meeting, Roskam brings pro-gun rights teens to talk gun violence prevention with students

\tabitem{Model erroneously predicted \say{2nd Amendment/Gun Rights}: \say{\textbf{pro-gun}}, \say{\textbf{rights}}}

\textit{\underline{\textbf{+ API tag:}}} Peter Roskam Republican Party Democratic Party United States Congress Illinois Member of Congress 2018 United States elections United States House of Representatives Presidency of Donald Trump House Committee on Ways and Means

\tabitem{Model correctly predicted \say{Politics}: using lexical cues of politician and party names and political terms: \textbf{\say{Trump}}, \textbf{\say{Republican}}, \textbf{\say{Democratic}}, \textbf{\say{Congress}}}}\\
\midrule
\tiny{Overgeneralizing highly-correlated features along with (long-distance dependencies)} & \tiny{Highly correlated words and phrases that do not directly cue frames, and are used in different contexts.}&
\tiny{\textit{\underline{\textbf{Example Frame:} Race/Ethnicity}}

\textit{\underline{\textbf{Headline:}}} Lawyers call US gun charges for Mexican man \textbf{\say{vindictive}}

\tabitem{Model erroneously predicted \say{Mental Health}: \textbf{\say{vindictive}}}

\textit{\underline{\textbf{+ API tag:}} }Shooting of Kate Steinle Acquittal Murder San Francisco Homicide Jury Death Defendant Illegal immigration Manslaughter

\tabitem{Model correctly predicted \say{Race/Ethnicity}: better context}}\\
\bottomrule
\end{tabular}
\caption{\small{Framing classification common error types, their definition and examples indicating the prediction error and how additional features in our top-performing methods of \textsc{BERT Headline + API}
can drive correct predictions.}}
\label{table:frame_errors}
\end{table*}
One possible cause for why predicting or even deciding on image relevance is challenging (e.g., in our dataset roughly half of the lead images are irrelevant) may be related to the nature of frames and images of U.S. gun violence coverage. Although frames or perspectives are abstract concepts, some frames such as \say{Society/Culture}, which focuses on \textit{society-wide factors} related to gun violence\footnote{Definitions of gun violence frames taken from the publicly available GVFC codebook and  dataset \url{https://derrywijaya.github.io/GVFC.html}}, are by nature more abstract and thus harder to convey through images than more concrete frames, such as \say{Politics}, which focuses on the political issues around guns and can be expressed more easily via images of politicians. 

As the ability of images to usefully represent a word is strongly dependent on how concrete or abstract the word is \cite{gilhooly1980age,friendly1982toronto}, a measure of frame concreteness or the ease of identifying tangible concepts and mental images that arise in correspondence to the frame, should  \textit{relate} to the ease of expressing frames via images or the ratio of relevant images for frames (Table~\ref{tab:numsamples}). 

To measure concreteness of frames and test this hypothesis, we trained a regression network that takes as input a word's vector representation as extracted and concatenated from the last 4 layers of the pre-trained BERT model and outputs its concreteness measure between 1 (most abstract) to 5 (most concrete). We use a dataset created by \citet{brysbaert2014concreteness}, which contains human evaluations of concreteness for 39,954 English words, to train and evaluate the network, achieving a high 0.95 Pearson's Correlation between our concreteness predictions and the ground-truth measures. The concreteness of a frame is then measured as the average concreteness of non named-entity words in its headlines (we treat named-entities as having a concreteness measure of 5). As seen in Figure \ref{fig:concreteness}, no frame has a high (> 4) average concreteness.

We observe, however, that some of the more concrete frames have higher ratios of their images annotated as relevant, e.g., \say{Politics}, which is the most concrete frame, has 65\% of its images annotated as relevant compared to just 10\% for \say{Society/Culture}, the least concrete frame. 

In fact, for most frames, a higher average concreteness implies a higher image relevance ratio. 
Exceptions to this are \say{Economic Consequences} and \say{Race/Ethnicity}. Although more words in the former are abstract (e.g., \textit{sales, demand, supply}), it is relatively easy to identify relevant images for economic consequence: e.g. \textit{company logos, gun stores}. On the other hand, although more words in \say{Race/Ethnicity} may be concrete: e.g., people/organization names, ethnic minority group names, or hate group names, it is harder to find relevant images for this frame in news articles. This may be due to editors in mainstream media, and photographers or journalists, withholding certain imagery from readers for fear of causing offence or shock, or for fear that a part of their audience may abandon the publication altogether \cite{times_violent_news}. Thus, although we find that frame concreteness is related to image relevance ratio (Pearson correlation of 0.69), there may be other factors that influence the choice of images for news articles that are beyond relevance to frames. 

We also report per frame classification F1 scores for the baseline and our best performing models on \textit{Articles with Relevant Images} and \textit{All Articles}, i.e., when using the article headline alone, or with the the API, the Caption, and the Summary 

in Figure~\ref{fig:per_frame}.
Performance when using information from articles and images is remarkable for frames with either high image relevance ratios or high concreteness. 

In \textit{Articles with Relevant Images}, the frames with the highest image relevance ratio i.e.,  \say{Politics} shows an impressive 96.6\% F1 score with the headline and the API tags, followed by \say{Public Opinion} (87.1\%); while frames with few relevant images (\say{2nd Amendment}, \say{Society/Culture}), have a substantially lower F1 scores. 
\begin{table*}[ht!]
\centering
\resizebox{\textwidth}{!}{
\begin{tabular}{M M M M M M M M} 
\toprule
\textbf{racial group}&\multicolumn{3}{c}{\textbf{headlines}} & \multicolumn{4}{c}{\textbf{images}}\\
\cmidrule(lr){2-4}\cmidrule(lr){5-8}
&\multicolumn{1}{c}{\textbf{all}}&\multicolumn{1}{c}{\textbf{P}}&\multicolumn{1}{c}{\textbf{PO}}&\multicolumn{1}{c}{\textbf{V}} &\multicolumn{1}{c}{\textbf{all}}&\multicolumn{1}{c}{\textbf{P}}&\multicolumn{1}{c}{\textbf{PO}}\\
\midrule
white &0.92&100&-&-&50.6&16&71\\
black & 2.15 & 7 & - & 93 & 4.2 &14&45\\
white \& black &0.69&-&- &-&2.7&3&84\\
asian &-&-&- &-&0.46&16&50\\
white \& asian &-&-&-&-&0.006&-&100\\
hispanic &0.07&100&-&-&-&-&-\\
jewish &7.1&2&- &98&-&-&-\\
other &89&-&- &-&-&-&-\\
\bottomrule
\end{tabular}}
\caption{\small In column \say{all} we see the percentage (\%) of headline mentions or image portrayals of certain racial groups in the 1,300 articles of the GVFC dataset. In columns \textbf{P}, \textbf{PO}, \textbf{V}, we can see the percentage (\%) of the people in each of these groups who are either the \textbf{P}erpetrator, \textbf{PO}litician (or Public Figure), or \textbf{V}ictim.}
\label{tab:minority_stats}
\end{table*}

On \textit{All Articles}, the inclusion of articles with irrelevant images can hurt performance for frames with high image relevance ratio such as Politics and Public Opinion. However, other frames may benefit from having more examples to learn from. For example, we observe that a low image relevance but a highly concrete frame such as \say{Race/Ethnicity} benefits significantly from this inclusion, reaching a high F1 score of 90.6\% using \textit{All Articles} from 82.1\% using only \textit{Articles with Relevant Images} as the model learns more lexical cues, e.g., named-entities from headlines of more articles, including those with irrelevant images. 

Concreteness may also augment relevance in explaining improved performance for some frames on \textit{All Articles}. We observe high correlations between frame average concreteness and average F1 scores on articles with relevant images (Pearson correlation of 0.93) and on \textit{All Articles} (Pearson correlation of 0.94), which exceed correlations between frame image relevance ratio and average F1 scores (Pearson correlation of 0.81 and 0.67 on articles with relevant images and on \textit{All Articles}, respectively). These findings suggest that concreteness might be worth exploring for frame prediction and use of imagery in the future, in addition to concreteness annotation in framing datasets. 

To complete our analysis, we applied the frame prediction error taxonomy proposed by \citet{mendelsohn2021modeling} to our news framing with image- and article-derived information, to identify and summarize common classification errors in Table~\ref{table:frame_errors}. We provide specific examples, highlight  possible error sources and observe how background information in \textsc{BERT Headline + API},
drives correct predictions, illustrating our previous remarks.

\section{Conclusions}

We presented the first ever study and dataset on computational multimodal framing. Our results show that image-derived contextual features can be useful for providing missing contextual or background information that can improve frame prediction significantly, particularly for concrete frames or frames with relevant images. We also proposed methods for predicting frame image relevance and for measuring frame concreteness, which we define as the ease of expressing frames via images. 

\section{Ethical Considerations}
Regarding the data we collected i.e., the Gun Violence Frame Corpus, we have made sure that there is no design experiment that was biased toward extracting only articles from a particular ethnic or minority group. We collect articles that had at least one keyword in their headlines from the
following list, based on previous literature on gun violence framing analysis as described in \citet{SLiu19}.
The keywords are {\say{gun}, \say{firearm}, \say{NRA}, \say{2nd amendment}, \say{second amendment}, \say{AR15}, \say{assault weapon}, \say{rifle}, \say{Brady act}, \say{Brady bill}, \say{mass shooting}}. The articles were retrieved in 2018 from 21 media outlets, from a list of top, in terms of website traffic, U.S. news websites;
and synthesizing these lists towards creating one list that contained news sites from the left, center, and right
sides of the ideological spectrum based on categories defined in \citet{mediacloud,PRC16,AdFontes}. 

Our analysis of the headlines and images gave each racial group's mentions' and portrayals' percentages provided in Table \ref{tab:minority_stats}. We notice that when racial groups are mentioned in news headlines (which is only in $\sim$11\% of all headlines), they are used to refer to victims of race-related gun violence incidents. Among the mentions, Blacks and Jews are the most common, as the corpus contains articles from 2018 that reported the mass shooting at Pittsburgh Synagogue\footnote{\url{https://en.wikipedia.org/wiki/Pittsburgh_synagogue_shooting}} and several high-profile shootings of black men--widely reported as instances of the controversial and race-related ``Stand Your Ground Law"\footnote{\url{https://en.wikipedia.org/wiki/Shooting_of_Markeis_McGlockton}}, all of which occurred in 2018. 

In terms of images, we notice that they are dominated by white people ($\sim$50\% of all images). However, the majority of them ($\sim$71\% of all white people images) are images of politicians or public figures related to gun laws/debates. There are much less images of victims and perpetrators (only $\sim$9\% of all images each). In terms of victims vs. perpetrators, there are more images of black victims ($\sim$1.8\%) than black perpetrators ($\sim$0.6\%). The same applies to Asian, while for whites the numbers of victim and perpetrator images are more balanced. 
Based on our data analysis, in which we saw different coverage in terms of racial groups in headlines vs. images, examining the difference between the race of the people mentioned in headlines and the race of those portrayed in the images would be an interesting future research direction. 

\section*{Acknowledgments}
\vspace{-1mm}
This work is supported in part by the U.S. NSF grant 1838193, DARPA HR001118S0044 (the LwLL program), and the Department of the Air Force FA8750-19-2-3334 (Semi-supervised Learning of Multimodal Representations). The U.S. Government is authorized to reproduce and distribute
reprints for Governmental purposes. The views and conclusions contained in this publication are those of the authors and should not be interpreted as representing official policies or endorsements of DARPA, the Air Force, and the U.S. Government.

\bibliography{anthology,custom}

\begin{thebibliography}{51}
\expandafter\ifx\csname natexlab\endcsname\relax\def\natexlab#1{#1}\fi

\bibitem[{{Ad Fontes Media}(2019)}]{AdFontes}
{Ad Fontes Media}. 2019.
\newblock \href {https://www.adfontesmedia.com/intro-to-the-media-bias-chart/} {The media bias chart}.

\bibitem[{Aky{\"u}rek et~al.(2020)Aky{\"u}rek, Guo, Elanwar, Ishwar, Betke, and Wijaya}]{akyurek-etal-2020-multi}
Afra~Feyza Aky{\"u}rek, Lei Guo, Randa Elanwar, Prakash Ishwar, Margrit Betke, and Derry~Tanti Wijaya. 2020.
\newblock \href {https://doi.org/10.18653/v1/2020.acl-main.763} {Multi-label and multilingual news framing analysis}.
\newblock In \emph{Proceedings of the 58th Annual Meeting of the Association for Computational Linguistics}, pages 8614--8624, Online. Association for Computational Linguistics.

\bibitem[{Brysbaert et~al.(2014)Brysbaert, Warriner, and Kuperman}]{brysbaert2014concreteness}
Marc Brysbaert, Amy~Beth Warriner, and Victor Kuperman. 2014.
\newblock \href {https://link.springer.com/article/10.3758/s13428-013-0403-5} {Concreteness ratings for 40 thousand generally known english word lemmas}.
\newblock \emph{Behavior research methods}, 46(3):904--911.

\bibitem[{Burns et~al.(2020)Burns, Kim, Wijaya, Saenko, and Plummer}]{burns2020learning}
Andrea Burns, Donghyun Kim, Derry Wijaya, Kate Saenko, and Bryan~A Plummer. 2020.
\newblock \href {https://arxiv.org/pdf/2004.04312.pdf} {Learning to scale multilingual representations for vision-language tasks}.
\newblock In \emph{European Conference on Computer Vision}, pages 197--213. Springer.

\bibitem[{Caglayan et~al.(2019)Caglayan, Madhyastha, Specia, and Barrault}]{caglayan2019probing}
Ozan Caglayan, Pranava Madhyastha, Lucia Specia, and Lo{\"\i}c Barrault. 2019.
\newblock \href {https://aclanthology.org/N19-1422/} {Probing the need for visual context in multimodal machine translation}.
\newblock \emph{arXiv preprint arXiv:1903.08678}.

\bibitem[{Caple(2010)}]{Caple10}
Helen Caple. 2010.
\newblock \href {http://www.hamptonpress.com/Merchant2/merchant.mvc?Screen=PROD&Product_Code=1-57273-93833} {What you see and what you get: {T}he evolving role of news photographs in an {A}ustralian broadsheet}.
\newblock \emph{Journalism and Meaning-Making: Reading the Newspaper}, pages 199--220.

\bibitem[{Card et~al.(2015)Card, Boydstun, Gross, Resnik, and Smith}]{CardBoGrReSm15}
Dallas Card, Amber~E. Boydstun, Justin~H. Gross, Philip Resnik, and Noah~A. Smith. 2015.
\newblock \href {https://doi.org/10.3115/v1/P15-2072} {The media frames corpus: Annotations of frames across issues}.
\newblock In \emph{Proceedings of the 53rd Annual Meeting of the Association for Computational Linguistics and the 7th International Joint Conference on Natural Language Processing (Volume 2: Short Papers)}, pages 438--444, Beijing, China. Association for Computational Linguistics.

\bibitem[{Coleman and Wu(2015)}]{ColemanWu15}
Renita Coleman and Denis Wu. 2015.
\newblock \href {https://www.researchgate.net/publication/279929835_Image_and_emotion_in_voter_decisions_The_affect_agenda} {\emph{Image and emotion in voter decisions: {T}he affect agenda}}.
\newblock Lexington Books.

\bibitem[{Dan(2017)}]{Dan17}
Viorela Dan. 2017.
\newblock \href {https://www.routledge.com/Integrative-Framing-Analysis-Framing-Health-through-Words-and-Visuals/Dan/p/book/9780367889081} {\emph{Integrative framing analysis: {F}raming health through words and visuals}}.
\newblock Routledge.

\bibitem[{Dauphin et~al.(2017)Dauphin, Fan, Auli, and Grangier}]{dauphin2017language}
Yann~N. Dauphin, Angela Fan, Michael Auli, and David Grangier. 2017.
\newblock \href {http://proceedings.mlr.press/v70/dauphin17a.html} {Language modeling with gated convolutional networks}.
\newblock In \emph{Proceedings of the 34th International Conference on Machine Learning}, volume~70 of \emph{Proceedings of Machine Learning Research}, pages 933--941. PMLR.

\bibitem[{Deng et~al.(2009)Deng, Dong, Socher, Li, Li, and Fei-Fei}]{DengDoSoLiLiLi09}
Jia Deng, Wei Dong, Richard Socher, Li-Jia Li, Kai Li, and Li~Fei-Fei. 2009.
\newblock \href {https://ieeexplore.ieee.org/abstract/document/5206848?casa_token=pKe_jirPwC4AAAAA:Us5-T-j0XxBPI2M16FOc2r_3kwHiUAvBfB41Abr85xGJUZsEDY1pmir0UHS2A4to021bVGeL} {Image{N}et: {A} large-scale hierarchical image database}.
\newblock In \emph{IEEE Computer Vision and Pattern Recognition (CVPR)}, pages 248--255.

\bibitem[{Devlin et~al.(2018)Devlin, Chang, Lee, and Toutanova}]{DevlinChLeTo18}
Jacob Devlin, Ming{-}Wei Chang, Kenton Lee, and Kristina Toutanova. 2018.
\newblock \href {http://arxiv.org/abs/1810.04805} {{BERT:} {P}re-training of deep bidirectional transformers for language understanding}.
\newblock \emph{CoRR}, abs/1810.04805.

\bibitem[{Devlin et~al.(2019)Devlin, Chang, Lee, and Toutanova}]{devlin-etal-2019-bert}
Jacob Devlin, Ming-Wei Chang, Kenton Lee, and Kristina Toutanova. 2019.
\newblock \href {https://doi.org/10.18653/v1/N19-1423} {{BERT}: Pre-training of deep bidirectional transformers for language understanding}.
\newblock In \emph{Proceedings of the 2019 Conference of the North {A}merican Chapter of the Association for Computational Linguistics: Human Language Technologies, Volume 1 (Long and Short Papers)}, pages 4171--4186, Minneapolis, Minnesota. Association for Computational Linguistics.

\bibitem[{Dodge et~al.(2020)Dodge, Ilharco, Schwartz, Farhadi, Hajishirzi, and Smith}]{dodge2020fine}
Jesse Dodge, Gabriel Ilharco, Roy Schwartz, Ali Farhadi, Hannaneh Hajishirzi, and Noah Smith. 2020.
\newblock \href {https://arxiv.org/pdf/2002.06305.pdf} {Fine-tuning pretrained language models: Weight initializations, data orders, and early stopping}.
\newblock \emph{arXiv preprint arXiv:2002.06305}.

\bibitem[{Drakulich(2015)}]{drakulich2015explicit}
Kevin~M Drakulich. 2015.
\newblock \href {https://doi.org/10.1093/socpro/spu003} {Explicit and hidden racial bias in the framing of social problems}.
\newblock \emph{Social Problems}, 62(3):391--418.

\bibitem[{Entman(1993)}]{Entman93}
Robert~M Entman. 1993.
\newblock \href {https://onlinelibrary.wiley.com/doi/abs/10.1111/j.1460-2466.1993.tb01304.x?source=post_elevate_sequence_page---------------------------} {Framing: {T}oward clarification of a fractured paradigm}.
\newblock \emph{Journal of communication}, 43(4):51--58.

\bibitem[{Field et~al.(2018)Field, Kliger, Wintner, Pan, Jurafsky, and Tsvetkov}]{FieldKlWiPaJuTs18}
Anjalie Field, Doron Kliger, Shuly Wintner, Jennifer Pan, Dan Jurafsky, and Yulia Tsvetkov. 2018.
\newblock \href {https://aclanthology.info/papers/D18-1393/d18-1393} {Framing and agenda-setting in russian news: a computational analysis of intricate political strategies}.
\newblock In \emph{Proceedings of the 2018 Conference on Empirical Methods in Natural Language Processing, Brussels, Belgium, October 31 - November 4, 2018}, pages 3570--3580.

\bibitem[{Friendly et~al.(1982)Friendly, Franklin, Hoffman, and Rubin}]{friendly1982toronto}
Michael Friendly, Patricia~E Franklin, David Hoffman, and David~C Rubin. 1982.
\newblock \href {https://link.springer.com/content/pdf/10.3758%2FBF03203275.pdf} {The toronto word pool: Norms for imagery, concreteness, orthographic variables, and grammatical usage for 1,080 words}.
\newblock \emph{Behavior Research Methods \& Instrumentation}, 14(4):375--399.

\bibitem[{Gilhooly and Logie(1980)}]{gilhooly1980age}
Ken~J Gilhooly and Robert~H Logie. 1980.
\newblock \href {https://link.springer.com/content/pdf/10.3758%2FBF03201693.pdf} {Age-of-acquisition, imagery, concreteness, familiarity, and ambiguity measures for 1,944 words}.
\newblock \emph{Behavior research methods \& instrumentation}, 12(4):395--427.

\bibitem[{Guo et~al.(2021)Guo, Mays, Zhang, Wijaya, and Betke}]{guo2021makes}
Lei Guo, Kate Mays, Yiyan Zhang, Derry Wijaya, and Margrit Betke. 2021.
\newblock \href {https://doi.org/10.1080/15205436.2021.1898644} {What makes gun violence a (less) prominent issue? a computational analysis of compelling arguments and selective agenda setting}.
\newblock \emph{Mass communication and society}, pages 1--25.

\bibitem[{Hamborg et~al.(2019)Hamborg, Zhukova, and Gipp}]{HamborgZhGi19}
Felix Hamborg, Anastasia Zhukova, and Bela Gipp. 2019.
\newblock \href {https://link.springer.com/chapter/10.1007/978-3-030-15742-5_17} {Illegal aliens or undocumented immigrants? {T}owards the automated identification of bias by word choice and labeling}.
\newblock Technical report, University of Konstanz, Germany.

\bibitem[{He et~al.(2016)He, Zhang, Ren, and Sun}]{HeZhReSu16}
Kaiming He, Xiangyu Zhang, Shaoqing Ren, and Jian Sun. 2016.
\newblock \href {https://www.cv-foundation.org/openaccess/content_cvpr_2016/papers/He_Deep_Residual_Learning_CVPR_2016_paper.pdf} {Deep residual learning for image recognition}.
\newblock In \emph{IEEE Conference on Computer Vision and Pattern Recognition}, pages 770--778.

\bibitem[{Hewitt et~al.(2018)Hewitt, Ippolito, Callahan, Kriz, Wijaya, and Callison-Burch}]{hewitt2018learning}
John Hewitt, Daphne Ippolito, Brendan Callahan, Reno Kriz, Derry~Tanti Wijaya, and Chris Callison-Burch. 2018.
\newblock \href {https://aclanthology.org/P18-1239/} {Learning translations via images with a massively multilingual image dataset}.
\newblock In \emph{Proceedings of the 56th Annual Meeting of the Association for Computational Linguistics (Volume 1: Long Papers)}, pages 2566--2576.

\bibitem[{Khani et~al.(2021)Khani, Tourni, Rasooli, Callison-Burch, and Wijaya}]{khani2021cultural}
Nikzad Khani, Isidora Tourni, Mohammad~Sadegh Rasooli, Chris Callison-Burch, and Derry~Tanti Wijaya. 2021.
\newblock \href {https://aclanthology.org/2021.naacl-main.19/} {Cultural and geographical influences on image translatability of words across languages}.
\newblock In \emph{Proceedings of the 2021 Conference of the North American Chapter of the Association for Computational Linguistics: Human Language Technologies}, pages 198--209.

\bibitem[{Kim et~al.(2020)Kim, Saito, Saenko, Sclaroff, and Plummer}]{kim2020mule}
Donghyun Kim, Kuniaki Saito, Kate Saenko, Stan Sclaroff, and Bryan Plummer. 2020.
\newblock \href {https://arxiv.org/abs/1909.03493} {Mule: Multimodal universal language embedding}.
\newblock In \emph{Proceedings of the AAAI Conference on Artificial Intelligence}, volume~34, pages 11254--11261.

\bibitem[{Krippendorff(2018)}]{krippendorff2018content}
Klaus Krippendorff. 2018.
\newblock \href {https://us.sagepub.com/en-us/nam/content-analysis/book258450} {\emph{Content analysis: An introduction to its methodology}}.
\newblock Sage publications.

\bibitem[{Lin et~al.(2017)Lin, Goyal, Girshick, He, and Doll{\'a}r}]{lin2017focal}
Tsung-Yi Lin, Priya Goyal, Ross Girshick, Kaiming He, and Piotr Doll{\'a}r. 2017.
\newblock \href {https://openaccess.thecvf.com/content_ICCV_2017/papers/Lin_Focal_Loss_for_ICCV_2017_paper.pdf} {Focal loss for dense object detection}.
\newblock In \emph{IEEE International Conference on Computer Vision}, pages 2980--2988.

\bibitem[{Liu et~al.(2019)Liu, Guo, Mays, Betke, and Wijaya}]{SLiu19}
Siyi Liu, Lei Guo, Kate Mays, Margrit Betke, and Derry~Tanti Wijaya. 2019.
\newblock \href {https://doi.org/10.18653/v1/K19-1047} {Detecting frames in news headlines and its application to analyzing news framing trends surrounding {U}.{S}. gun violence}.
\newblock In \emph{Proceedings of the 23rd Conference on Computational Natural Language Learning (CoNLL)}, pages 504--514, Hong Kong, China. Association for Computational Linguistics.

\bibitem[{Liu and Lapata(2019)}]{lapata-2019-text}
Yang Liu and Mirella Lapata. 2019.
\newblock \href {"https://www.aclweb.org/anthology/D19-1387"} {Text summarization with pretrained encoders}.
\newblock \emph{arXiv preprint arXiv:1908.08345}.

\bibitem[{Liu et~al.(2020)Liu, Ott, Goyal, Du, Joshi, Chen, Levy, Lewis, Zettlemoyer, and Stoyanov}]{liu2019roberta}
Yinhan Liu, Myle Ott, Naman Goyal, Jingfei Du, Mandar Joshi, Danqi Chen, Omer Levy, Mike Lewis, Luke Zettlemoyer, and Veselin Stoyanov. 2020.
\newblock \href {https://openreview.net/forum?id=SyxS0T4tvS} {Ro{\{}bert{\}}a: A robustly optimized {\{}bert{\}} pretraining approach}.

\bibitem[{Lowrey(2002)}]{Lowrey02}
Wilson Lowrey. 2002.
\newblock \href {https://doi.org/10.1207/S15327825MCS0504_03} {Word people vs. picture people: {N}ormative differences and strategies for control over work among newsroom subgroups}.
\newblock \emph{Mass Communication \& Society}, 5(4):411--432.

\bibitem[{MediaCloud()}]{mediacloud}
MediaCloud. 2018.
\newblock \href {https://mediacloud.org} {https://mediacloud.org}.

\bibitem[{Mendelsohn et~al.(2021)Mendelsohn, Budak, and Jurgens}]{mendelsohn2021modeling}
Julia Mendelsohn, Ceren Budak, and David Jurgens. 2021.
\newblock \href {https://arxiv.org/pdf/2104.06443.pdf} {Modeling framing in immigration discourse on social media}.
\newblock \emph{arXiv preprint arXiv:2104.06443}.

\bibitem[{Messaris and Abraham(2001)}]{MessarisAb01}
Paul Messaris and Linus Abraham. 2001.
\newblock \href {https://www.taylorfrancis.com/chapters/edit/10.4324/9781410605689-22/role-images-framing-news-stories-paul-messaris-linus-abraham} {The role of images in framing news stories}.
\newblock In \emph{Framing public life}, pages 231--242. Routledge.

\bibitem[{Mosbach et~al.(2021)Mosbach, Andriushchenko, and Klakow}]{mosbach2021on}
Marius Mosbach, Maksym Andriushchenko, and Dietrich Klakow. 2021.
\newblock \href {https://openreview.net/forum?id=nzpLWnVAyah} {On the stability of fine-tuning {\{}bert{\}}: Misconceptions, explanations, and strong baselines}.
\newblock In \emph{International Conference on Learning Representations}.

\bibitem[{Naderi and Hirst(2017)}]{NaderiHi17}
Nona Naderi and Graeme Hirst. 2017.
\newblock \href {https://doi.org/10.26615/978-954-452-049-6\_070} {Classifying frames at the sentence level in news articles}.
\newblock In \emph{Proceedings of the International Conference Recent Advances in Natural Language Processing, {RANLP} 2017, Varna, Bulgaria, September 2 - 8, 2017}, pages 536--542. {INCOMA} Ltd.

\bibitem[{Nguyen et~al.(2013)Nguyen, Ying, and Resnik}]{NguyenBoRe13}
Viet-An Nguyen, Jordan~L Ying, and Philip Resnik. 2013.
\newblock \href {https://proceedings.neurips.cc/paper/2013/file/f5deaeeae1538fb6c45901d524ee2f98-Paper.pdf} {Lexical and hierarchical topic regression}.
\newblock In \emph{Advances in Neural Information Processing Systems}, volume~26. Curran Associates, Inc.

\bibitem[{Paivio et~al.(1968)Paivio, Yuille, and Madigan}]{paivio1968concreteness}
Allan Paivio, John~C Yuille, and Stephen~A Madigan. 1968.
\newblock \href {https://doi.org/10.1037/h0025327} {Concreteness, imagery, and meaningfulness values for 925 nouns.}
\newblock \emph{Journal of experimental psychology}, 76(1p2):1.

\bibitem[{{Pew Research Center}(2016)}]{PRC16}
{Pew Research Center}. 2016.
\newblock \href {https://www.pewresearch.org/pj_14-10-21_mediapolarization-08-2/} {Ideological placement of each source’s audience}.

\bibitem[{Powell et~al.(2015)Powell, Boomgaarden, De~Swert, and de~Vreese}]{PowellBoDeDe15}
Thomas~E Powell, Hajo~G Boomgaarden, Knut De~Swert, and Claes~H de~Vreese. 2015.
\newblock \href {https://onlinelibrary.wiley.com/doi/abs/10.1111/jcom.12184} {A clearer picture: {T}he contribution of visuals and text to framing effects}.
\newblock \emph{Journal of Communication}, 65(6):997--1017.

\bibitem[{Rasooli et~al.(2021)Rasooli, Callison-Burch, and Wijaya}]{rasooli2021wikily}
Mohammad~Sadegh Rasooli, Chris Callison-Burch, and Derry~Tanti Wijaya. 2021.
\newblock \href {https://arxiv.org/abs/2104.08384v1} {" wikily" neural machine translation tailored to cross-lingual tasks}.
\newblock \emph{arXiv preprint arXiv:2104.08384}.

\bibitem[{Redmon and Farhadi(2018)}]{farhadi2018yolov3}
Joseph Redmon and Ali Farhadi. 2018.
\newblock \href {http://arxiv.org/abs/1804.02767} {Yolov3: An incremental improvement}.
\newblock \emph{CoRR}, abs/1804.02767.

\bibitem[{Reese et~al.(2001)Reese, Gandy~Jr, Grant et~al.}]{ReeseGaGr01}
Stephen~D Reese, Oscar~H Gandy~Jr, August~E Grant, et~al. 2001.
\newblock \href {https://www.routledge.com/Framing-Public-Life-Perspectives-on-Media-and-Our-Understanding-of-the/Reese-Gandy-Jr-Grant/p/book/9780805849264} {\emph{Framing public life: {P}erspectives on media and our understanding of the social world}}.
\newblock Routledge.

\bibitem[{Ritchin(2014)}]{times_violent_news}
Fred Ritchin. 2014.
\newblock \href {https://time.com/3705884/why-violent-news-images-matter//} {Why violent news images matter}.
\newblock [time.com;4-September-2014].

\bibitem[{Specia et~al.(2016)Specia, Frank, Sima’An, and Elliott}]{specia2016shared}
Lucia Specia, Stella Frank, Khalil Sima’An, and Desmond Elliott. 2016.
\newblock \href {https://aclanthology.org/W16-2346/} {A shared task on multimodal machine translation and crosslingual image description}.
\newblock In \emph{Proceedings of the First Conference on Machine Translation: Volume 2, Shared Task Papers}, pages 543--553.

\bibitem[{Tran et~al.(2020)Tran, Mathews, and Xie}]{tran2020transform}
Alasdair Tran, Alexander Mathews, and Lexing Xie. 2020.
\newblock \href {https://openaccess.thecvf.com/content_CVPR_2020/papers/Tran_Transform_and_Tell_Entity-Aware_News_Image_Captioning_CVPR_2020_paper.pdf} {Transform and tell: Entity-aware news image captioning}.
\newblock In \emph{Proceedings of the IEEE/CVF Conference on Computer Vision and Pattern Recognition}, pages 13035--13045.

\bibitem[{Vaswani et~al.(2017)Vaswani, Shazeer, Parmar, Uszkoreit, Jones, Gomez, Kaiser, and Polosukhin}]{vaswani2017attention}
Ashish Vaswani, Noam Shazeer, Niki Parmar, Jakob Uszkoreit, Llion Jones, Aidan~N Gomez, \L~ukasz Kaiser, and Illia Polosukhin. 2017.
\newblock \href {https://proceedings.neurips.cc/paper/2017/file/3f5ee243547dee91fbd053c1c4a845aa-Paper.pdf} {Attention is all you need}.
\newblock In \emph{Advances in Neural Information Processing Systems}, volume~30. Curran Associates, Inc.

\bibitem[{Wang et~al.(2020)Wang, Tran, and Feiszli}]{wang2020makes}
Weiyao Wang, Du~Tran, and Matt Feiszli. 2020.
\newblock \href {https://openaccess.thecvf.com/content_CVPR_2020/papers/Wang_What_Makes_Training_Multi-Modal_Classification_Networks_Hard_CVPR_2020_paper.pdf} {What makes training multi-modal classification networks hard?}
\newblock In \emph{Proceedings of the IEEE/CVF Conference on Computer Vision and Pattern Recognition}, pages 12695--12705.

\bibitem[{Wessler et~al.(2016)Wessler, Wozniak, Hofer, and Lück}]{WesslerWoHoLu16}
Hartmut Wessler, Antal Wozniak, Lutz Hofer, and Julia Lück. 2016.
\newblock \href {https://doi.org/10.1177/1940161216661848} {Global multimodal news frames on climate change: A comparison of five democracies around the world}.
\newblock \emph{The International Journal of Press/Politics}, 21(4):423--445.

\bibitem[{Yao and Wan(2020)}]{yao2020multimodal}
Shaowei Yao and Xiaojun Wan. 2020.
\newblock \href {https://aclanthology.org/2020.acl-main.400/} {Multimodal transformer for multimodal machine translation}.
\newblock In \emph{Proceedings of the 58th Annual Meeting of the Association for Computational Linguistics}, pages 4346--4350.

\bibitem[{Zhang et~al.(2016)Zhang, Zhang, Li, and Qiao}]{zhang2016joint}
Kaipeng Zhang, Zhanpeng Zhang, Zhifeng Li, and Yu~Qiao. 2016.
\newblock \href {https://ieeexplore.ieee.org/abstract/document/7553523/?casa_token=B6t8wvMa_XAAAAAA:-n3C5OE5nW4zSgRrK_DntcFHW8NDgVl2ty35t_5oLHUxeRkeCumxoGEnQjV1uxvUFIwjBmaK} {Joint face detection and alignment using multitask cascaded convolutional networks}.
\newblock \emph{IEEE Signal Processing Letters}, 23(10):1499--1503.

\end{thebibliography}
\bibliographystyle{acl_natbib}

\clearpage
\appendix
\section{Appendix}
\label{appendix:a}
\footnotesize The first 16 entries in the \textbf{SRE} feature vector indicate the central \textbf{S}ubject of the image, with each Subject implying certain frame(s):
\\
\begin{enumerate}[labelsep=0.5cm,font=\footnotesize,noitemsep,nolistsep]
\itemsep0em  
\item \footnotesize People: Gun shooter/suspect (\textbf{\underline{Mental Health}})
\item \footnotesize People: Gun hobbyist/activist + gun-related activities with a hand (\textbf{\underline{Gun Rights}})
\item \footnotesize People: Victim/affected family and friends/bystanders (\textbf{\underline{Public Opinion}})
\item \footnotesize People: Politicians (\textbf{\underline{Politics}})
\item \footnotesize People: Law enforcement (e.g., police offers, security guards) (\textbf{\underline{Public Safety}})
\item \footnotesize Object: Firearm/bullets  (can mean anything or \textbf{\underline{Gun Control}} for certain gun images)
\item \footnotesize Object: Gun /hunting gear stores/gun show (\textbf{\underline{Economic Consequences}} or \textbf{\underline{Gun Rights}})
\item \footnotesize People: Demonstrators/Demonstrations (\textbf{\underline{Public Opinion})}
\item \footnotesize Object: Protest signs (\textbf{\underline{Gun Control}} or \textbf{\underline{Gun Rights}})
\item \footnotesize People/mainly object: Memorials objects and people (\textbf{\underline{Public Opinion}})
\item \footnotesize Object/people: Crime scene/police cars/people during or right after the crisis (episodic frame)
\item \footnotesize Object: Legislative buildings/courthouses (\textbf{\underline{Gun Control}} or \textbf{\underline{Politics}})
\item \footnotesize Object/people: School/campus/students indicating school/campus (\textbf{\underline{Public Safety}})
\item \footnotesize NRA objects/NRA representatives (\textbf{\underline{Gun Rights}} or \textbf{\underline{Economic Consequences}})
\item \footnotesize Object: Company buildings/logos (\textbf{\underline{Economic Consequence}})
\item \footnotesize Other 
\end{enumerate}

\vspace{3mm}
\footnotesize The last three entries in the feature vector indicate relevance to \textbf{R}ace or \textbf{E}thnicity:
\\
\begin{enumerate}[labelsep=0.5cm,font=\footnotesize,noitemsep,nolistsep]
\setcounter{enumi}{16}
\itemsep0em 
\item \footnotesize None 
\item \footnotesize Racial/ethnic minority groups /buildings of a specific group (\textbf{\underline{Ethnicity}}) (only if the central subject is from racial/ethnic minority group 
- not if there is only one or a couple of non-White people in a large crowd)  
\item \footnotesize KKK/white supremacy/hate groups (\textbf{\underline{Ethnicity}})
\end{enumerate}

\clearpage
\setlength\rotFPtop{242pt}
\begin{sidewaystable}[htbp]
\resizebox{\textwidth}{!}{
\begin{tabular}{l l l}
\toprule
\textbf{\Huge{Frame}} & \textbf{\Huge{Description}}& \textbf{\Huge{Examples}}\\
\midrule\\
\Huge{Politics} & \Huge{Political issues around guns and shootings, including:}&\Huge{ \say{Baltimore students walk out of class to protest gun violence}}\\&
\Huge{\tabitem Political campaigns and upcoming elections}
\\& \Huge{(e.g., using guns as a wedge issue or motivating force to get people to the polls)} \\&
\Huge{\tabitem Fighting between the Democratic and Republican parties, or politicians} \\&
\Huge{\tabitem Political money contributions from gun lobbies (e.g., NRA)} \\&
\Huge{\tabitem One political party or one politician’s stance on gun violence.}\\& 
\Huge{\tabitem Therefore, as long as the news headline mentions a politician’s name,}\\&
\Huge{it often indicates the theme of politics.} \\&
\Huge{\tabitem Often times, the politicians’ names or the party names should be mentioned.}\\&
\\
\Huge{Public Opinion} & \Huge{Public’s opinion}& \Huge{\say{How Illinois governor candidates would address gun violence}}\\&
\Huge{and the community’s reactions to gun-related issues, including:} &\Huge{\say{Trump warns Dems will 'take away your Second Amendment'}}\\& \Huge{(e.g., using guns as a wedge issue or motivating force to get people to the polls)} &\Huge{\say{Lindsey Graham: Both parties will suffer if Congress doesn't act on new gun bill}}\\&
\Huge{\tabitem Public opinion polls related to guns}\\& 
\Huge{\tabitem Protests} \\&
\Huge{\tabitem One political party or one politician’s stance on gun violence.}\\& 
\Huge{\tabitem Mourning victims of gun violence}\\&
\Huge{\tabitem The public’s emotional responses}\\&
\\
\Huge{Gun Control/Regulation} & \Huge{Issues related to regulating guns} 
&\Huge{\say{GOP lawmaker calls for age restriction on AR-15s}}\\& 
\Huge{through legislation and other institutional measures:}\\&
\Huge{\tabitem Enforcing and/or expanding background checks}&\Huge{\say{No bump stocks turned in to Denver police after ban}}\\& 
\Huge{\tabitem Limiting sale of guns and/or related dangerous equipment}\\& 
\Huge{(e.g., AR15s, semi-automatic rifles, bump stocks, Huge-capacity ammo)} \\&
\Huge{\tabitem Increasing age limits on gun purchases}\\& 
\Huge{\tabitem Implementing licensing and gun safety training programs} \\&
\\
\Huge{School/Public Space Safety} & \Huge{Issues related to institutional and school safety, including:}
&\Huge{\say{Preschoolers among students required to carry clear backpacks in Texas school district}}\\& 
\Huge{\tabitem Awareness and monitoring of \say{troubled} individuals by law enforcement (e.g., local police, FBI)} &\Huge{\say{Scott wants armed police at Stoneman Douglas after disturbing incidents at Parkland school}}\\& 
\Huge{\tabitem Safety measures in schools to prevent or mitigate shootings} & \Huge{\say{Sales of bulletproof school supplies spike after Florida shooting}}\\&
\Huge{(e.g., police/safety officers in the school, armed teachers, metal detectors, clear backpacks)}\\&
\Huge{\tabitem Note that a headline simply mentioning \say{school shooting}}\\&
\Huge{does not necessarily mean it uses this safety measure frame.}\\&
\\
\Huge{Economic consequences} & \Huge{Financial losses or gains, or the costs involved in gun-related issues, including:}
&\Huge{\say{The NRA Is In Deep, Deep Financial Trouble}}\\& 
\Huge{\tabitem The actual sales of firearms} \\& 
\Huge{\tabitem The financial consequences of gun regulation}\\&
\Huge{(e.g., lost tax revenue, or gun manufacturing companies moving to a different state)}\\&
\Huge{\tabitem The financial state of gun-related lobbying groups (e.g., the NRA)} \\& 
\Huge{\tabitem Federal budget for gun-related programs} \\&
\\
\Huge{Race/Ethnicity} &\Huge{Gun issues related to certain ethnic group(s), including:}
&\Huge{\say{Illegal immigrant acquitted of Kate Steinle's murder faces judge on gun charges}}\\& 
\Huge{\tabitem Angry, isolated white men as primary perpetrators of domestic gun violence} & \Huge{\say{The disparities in how black and white men die in gun violence, state by state}} \\& 
\Huge{\tabitem Immigrants from Mexico bringing in guns from across the border}\\&
\Huge{\tabitem Muslim \say{terrorists}} \\& 
\Huge{\tabitem Gun violence in African American communities} \\&
\\
\Huge{Mental Health} & \Huge{Issues related to individuals’ mental illnesses or emotional well-being,} 
&\Huge{\say{Gun debate hits home for families dealing with myths about violence, mental illness}}\\& 
\Huge{or the mental health system as a whole, including:}\\&
\Huge{\tabitem Predicting and preventing mental health breakdowns} & \Huge{\say{Renewed Debate Over Gun Access, Mental Health}} \\& 
\Huge{\tabitem Treating mental illness} & \Huge{\say{Las Vegas gunman lost money, became unstable before shooting}}\\&
\Huge{\tabitem Creating measures to ensure mentally ill people do not have access to guns}\\& 
\Huge{\tabitem Descriptions of individuals’ behavioral / personality traits}\\&
\Huge{that indicate instability, impulsivity, anger, etc.}\\&
\\
\Huge{2nd Amendment/Gun Rights} & \Huge{Related to the Constitution, the second amendment}, 
&\Huge{\say{Membership, interest in gun rights groups soar in the weeks after the Florida high school shooting}}\\& 
\Huge{and protection of individual liberty and gun ownership as a right, including:}\\&
\Huge{\tabitem Meaning of the 2nd amendment}  & \Huge{\say{Rapper ‘Killer Mike,’ NRA host Colion Noir: No guns would turn people into slaves}} \\& 
\Huge{\tabitem The irrefutability of one’s right to own guns} \\&
\Huge{\tabitem Gun ownership as critical to democracy and protecting oneself}\\&
\\
\Huge{Society/Culture} & \Huge{Societal-wide factors that are related to gun violence, including:}
&\Huge{\say{There's Not A Single Ounce Of Evidence To Link Mass Shootings To Video Games}}\\& 
\Huge{\tabitem Violence in media (e.g., TV/movies and video games)} \\& 
\Huge{\tabitem Social pressures that may incite someone to violence (e.g., cliques/bullying and isolation)} \\&
\Huge{\tabitem Breakdown in family structures, so there is a lack of familial support and stability}\\&
\Huge{\tabitem Breakdown in community structures (e.g., religious organizations, other civic-oriented groups)},\\&
\Huge{so there is a lack of community support and stability}\\&
\\\\
\bottomrule
\end{tabular}
}
\caption{News frames' description and Headline examples.}
\label{table:frame_analysis}
\end{sidewaystable}

\end{document}